# Behavioural Repertoire via Generative Adversarial Policy Networks

Marija Jegorova[1], Stéphane Doncieux[2], and Timothy Hospedales[1]

*Abstract*— **Learning algorithms are enabling robots to solve increasingly challenging real-world tasks. These approaches often rely on demonstrations and reproduce the behavior shown. Unexpected changes in the environment may require using different behaviors to achieve the same effect, for instance to reach and grasp an object in changing clutter. An emerging paradigm addressing this robustness issue is to learn a diverse set of successful behaviors for a given task, from which a robot can select the most suitable policy when faced with a new environment. In this paper, we explore a novel realization of this vision by learning a generative model over policies. Rather than learning a single policy, or a small fixed repertoire, our generative model for policies compactly encodes an unbounded number of policies and allows novel controller variants to be sampled. Leveraging our generative policy network, a robot can sample novel behaviors until it finds one that works for a new environment. We demonstrate this idea with an application of robust ball-throwing in the presence of obstacles. We show that this approach achieves a greater diversity of behaviors than an existing evolutionary approach, while maintaining good efficacy of sampled behaviors, allowing a Baxter robot to hit targets more often when ball throwing in the presence of obstacles.**

## I. Introduction

Robots are increasingly able to solve challenging tasks by learning controllers. While reinforcement or imitation learning approaches can be effective, they typically learn a single ideal solution to a given control problem, and the robustness of that solution to challenging situational variants (e.g., changing obstacles, or damage to the robot) is hard to guarantee. If a control policy fails due such an unexpected environmental change, robots can try to adapt their control policy to a new situation through re-planning [1] or adapting a learned policy [2]. Beyond such adaptation, when animals face a challenging environment in which a previously learned behavior fails, they also draw on an additional capability: leveraging a suite of other known behaviors that are expected to solve the task at hand [3]. Exploration within a set of diverse historical behaviors that solved a task can quickly lead to a solution that succeeds in a new environment [3]. Such behavioral repertoire-based approaches are emerging as promising techniques for robustly solving tasks [3], [4], [5].

Existing realizations of this robustness-through-diversity vision are often based on evolutionary algorithms that train a diverse set (population) of controllers that solve a given task [4], [5]. However this approach has several drawbacks: storing a large database of controllers is not compact, and there is only as much diversity as is contained in the population of controllers. We argue that a preferable instantiation of this vision is to learn a generative model over controllers. Firstly, it is compact – only the parameters of the generative model rather than a large list of controllers need to be stored. Secondly, the available diversity is not limited to the instances in a fixed length list. By sampling a generative model over controllers, an unlimited number of distinct controllers can be obtained. And with a sufficiently flexible generative model, sampled controllers need not be simple interpolations between controllers used to train the generative model. Samples could encode novel solutions to the problem by drawing diverse aspects of multiple training policies.

This approach is coherent with the exploratory behaviour of infants (and other animals) – specifically their ability to perform a behaviour in high variation so there is a distribution of actions associated with each behaviour [6]. Our method models this distribution. Following the example of other progressive sequential architectures – [7], [8], we propose a simple two-staged developmental framework where one first builds up the initial repertoire of actions (using methods such as quality-diversity search [9]), and then generalizes beyond this repertoire via our proposed generative model. Our progression from a library-based approach to a generative-model can also be considered a representational re-description [10], between developmental waves [6].

While conceptually appealing, training generative models over policies is non-trivial. The space of reasonable policies likely to solve a given task is a complicated manifold within the space of all policies, considering actuator redundancy, non-linearities and so on. We therefore propose to apply generative adversarial networks (GANs) [11] to model the distribution over policies that solve a given task using a neural network, thus defining a generative policy network (GPN). In our framework the GPN models the distribution over policy parameters, so that each sample from the GPN defines a specific robot controller. Multiple samples from the GPN therefore correspond to different solutions to the task that the GPN is trained on. To generate training data for the GPN we exploit quality-diversity (QD) search evolutionary algorithms [9] to find a diverse set of policies that solve a task. Once trained, a GPN then provides a compact source of diverse and novel policies likely to solve variants of that task. Compared to a conventional GAN, we find it beneficial to regularize GPN-training by requiring it to generate not only a controller but the outcome of running that controller (i.e. to simulate the forward model, or reconstruct the input goal state), and this is also useful as a way to pick promising policies (e.g., sample the GPN until a policy is drawn which is expected to work in the current environment).

We demonstrate our approach through the specific ap-plication of target-conditional ball-throwing [12], [13] in the presence of confounding obstacles. Throwing is often formalized as a contextual policy problem where a movement primitive for throwing is synthesized conditionally on the desired target position [12], [13]. In the presence of obstacles however, the most 'natural' way to throw to a given target may be blocked. Nevertheless, there are multiple throwing movements that hit a given target. We show that the ability to model and sample from a distribution of controllers allows the robot to find throwing controllers that can avoid any given obstacle.

[1] School of Informatics, University of Edinburgh, UK. {m.jegorova@sms.,t.hospedales@}ed.ac.uk. [2] Sorbonne Université, CNRS, ISIR, Paris, France. stephane.doncieux@upmc.fr.

## II. Related Work

*Learning Robot Control* Typical approaches to learning robot control include learning by demonstration [14], reinforcement learning to maximize some extrinsic reward [12], or demonstration-based initialization followed by policy search-based reinforcement learning to fine-tune the demonstrated policy. Those policy search algorithms in turn can often be categorized into gradient-based [15], [16] and gradient-free methods such as Bayesian optimization [17] and evolutionary search [18]. An advantage of evolutionary methods is that they often provide a population of policies as a byproduct, rather than a single best controller.

Where obstacles can impede behaviour, a standard robotics approach is to localise the obstacle and plan a movement that avoids it [19]. However this requires both (i) accurate 3D obstacle localisation and (ii) appropriate adaptive planning capabilities. One or other of these sensing and reasoning capabilities may not be available at the required efficacy level at a given developmental stage in an animal or robot. In contrast generating diverse behaviours and exploring them until one works has lower prerequisites and hence is suitable for earlier developmental stages.

*Behavioral Diversity in Robot Control* For a robot to be able to deal rapidly with new and unanticipated situations, a recently proposed approach consists of building a large repertoire of behaviors in which it should be possible to find one adapted to a newly arising situation or environment. The repertoire creation step can be done in a preliminary phase and a learned repertoire subsequently used to accelerate the adaptation to an unanticipated situation by relying on a selection process instead of a full learning process [3]. Promoting behavioral diversity is a key feature of a repertoire creation process. Driven by research on novelty search [20], evolutionary approaches have been adapted to generate a behaviorally diverse set of solutions instead of converging to a single solution optimizing a given fitness function [4]. These algorithms are called Quality Diversity algorithms [21], [22] and are used here to bootstrap the proposed method. Our proposed GPN builds on QD-search by leveraging its results as training data. However, in contrast to the selection-from-repertoire paradigm of QD, it has several interrelated benefits: (i) We can more compactly store a large repertoire by storing instead the parameters of a generative model that represents that repertoire of behaviors. (ii) Rather than a fixed size database of behaviors, the generative model can continue to sample unlimited new behaviors until a suitable one is found. (iii) Samples drawn from the generative model of behaviors can discover novelty beyond the initial training repertoire, by combining aspects from different training behaviors. (iv) Importantly the GPN approach is better suited for contextual policies. To solve a contextual policy task like diverse throwing to different targets, a library-based approach increases the required data collection and repertoire storage size dramatically because it would need to keep samples of many different throwing targets, and for each of those targets, samples of many different ways to throw there. In contrast, given a few samples of different throwing targets, a contextual policy GPN can extrapolate and draw many different controllers for throwing to any given target.

*Generating Diverse Policies* Another somewhat related work [23] covers diverse policy generation in a model-based framework. DIAYN [23] learns diverse skills (policies), assessing the diversity by the variety of states they visit in the process of RL-style unsupervised exploration. The main difference is that DIYAN tries to learn a small set of very distinct skills. While our GPN focuses on one skill type, but learns an infinite smoothly-varying manifold of controllers covering both all the potential goals (e.g., movement targets) and ways to achieve those goals. DIYAN also focuses more on initial exploration (and is thus analogous to QD-search in our pipeline), while we focus on compactly representing and exploiting the results of such an exploration process.

*Generative Adversarial Networks* Generative Adversarial Networks were proposed [11] to address the challenge of learning a neural network-based generative model for complex high-dimensional data. The key idea being that generator training is enabled by a second discriminator network that is simultaneously adversarially trained to distinguish true training data and the generator's synthetic examples. To improve its ability to fool the discriminator the generator must generate increasingly realistic synthetic samples. There have since been numerous extensions including convolutional GANs [24], conditional GANs [25], [26], and improvements of GAN training stability with regards to challenges such as non-convergence and mode-collapse [27], [28].

*Generative Adversarial Network Applications* The vast majority of GAN applications are in image generation tasks [11], [24], [29], [27], [26]. In robotics, GANs have been applied in robot haptic recognition [30]. Autoencoding VAE-GAN has been used for visual representation learning to process visual input in support of vision-based actuation in control [31]. GANs have also been referenced in an inverse reinforcement learning context [32] where an analogy is drawn between the distribution of state-action pairs encountered by an expert and the real data which should be matched by the distribution of state-action pairs encountered by a student policy and the synthetic data produced by a generator. To our knowledge neural network generators have not previously been applied to sample policies, or to the generation of diverse robot behaviors, as we explore here.

## III. METHOD

### A. Background: Generative Adversarial Networks

Generative adversarial networks are neural network generative models for complex high-dimensional data. In this framework, a generator G is trained to produce samples representative of a training data distribution $p_{data}(\mathbf{x})$. G takes as input a random noise vector $\mathbf{z}$, and for a given noise distribution $p(\mathbf{z})$, samples $\mathbf{x} = G(\mathbf{z})$, $\mathbf{z} \sim p(\mathbf{z})$ should follow the same distribution as the observed data $\mathbf{x} \sim p_{data}(\mathbf{x})$. Such generative neural networks are challenging to train, but [11] showed that they can be trained via a min-max game between the generator and an adversary (the discriminator) D:

$$\min_G \max_D V(G,D) = E_{\mathbf{x}\sim p_{data}(\mathbf{x})}[\log D(\mathbf{x})] + E_{\mathbf{z}\sim p(\mathbf{z})}[\log(1 - D(G(\mathbf{z})))] \tag{1}$$

where $\mathbf{x}$ stands for a data example, $\mathbf{z}$ a random noise vector, $D(\mathbf{x})$ represents the discriminator's estimate of the probability that $\mathbf{x}$ came from real data rather than the generator. and $D(G(\mathbf{z}))$ - a probability that data came from a generator. GANs can also be extended to model the distribution of data conditional on some observed context vector [24], in which case both the generator and the discriminator also take the conditioning data $\mathbf{c}$ as input. GANs are most commonly applied to generate images (e.g., $\mathbf{x}$ is a person image and $\mathbf{c}$ is the gender of that person). In the following we adapt them to generate policies $\mathbf{x}$ conditional on goals $\mathbf{c}$.

### B. Generative Policy Networks

*Unconditional Policies* Robot behaviors are defined by a control policy $\pi$ operating in some state space $\mathscr{S}$ and action space $\mathscr{A}$. Thus while generative models are conventionally used to define a distribution $p(\mathbf{x})$ over data instances $\mathbf{x}$, our GPN defines a distribution $p(\pi)$ over policies $\pi$, which are themselves functions $\pi : \mathscr{S} \to \mathscr{A}$. Given a set of training policies $D_{train} = \{\pi_i\}$, we train our GPN to estimate the distribution over observed policies. Assuming the policies in question lie in some parametric family, then each is identified by some parameter vector (e.g., weights in a neural network [33], radial basis function (RBF) kernels in a dynamic movement primitive (DMP) [14], [13]). By training a generator G to generate such parameters, samples from the generator are interpretable as controllers. In this case the discriminator enables the training of the generator by learning to distinguish between real policies in $D_{train}$ and generator synthesized policies $\pi = G(\mathbf{z})$. Once the generator learns to fool the discriminator, and assuming it does not mode collapse, then samples from the generator represent diverse control policies are novel yet statistically indistinguishable from the training policies. We denote sampling policies from the distribution implied by the generator $G(\mathbf{z})$ under a given noise distribution $p(\mathbf{z})$ as $\pi \sim p_G(\pi)$.

*Contextual Policies* We aim to go beyond simple fixed behaviors to work with contextual policies that are parameterized by a goal condition to achieve [12], [14]. For a goal directed policy such as our intended application of robot throwing, we need not just a controller (e.g., a throwing movement), but a conditioning mechanism that generates a controller that achieves the right goal (e.g., a throwing movement that hits a specific target). As described above, if the training set $D_{train}$ consists of controllers throwing to multiple different locations, then sampled policies $\pi \sim p_G(\pi)$ will throw to new locations within the distribution of training targets. If the training set $D_{train}$ consists of multiple controllers that throw in different ways to the same location, then $\pi \sim p_G(\pi)$ will sample novel policies that throw to that same location. In the contextual policy case we want a policy that achieves a specifiable goal. Thus we define a conditional generator $\pi = G(\mathbf{z},\mathbf{c})$, $\pi \sim p_{G(\mathbf{c})}(\pi)$ to sample policies $\pi$ that target a specific landing point $\mathbf{c}$. Thus we can both throw at a specified target (set the generator condition), and also find multiple ways to throw there (sample the generator).

### C. Application to Throwing

For the throwing application, we assume we are given a set of training policies, and we denote sampling these as $\pi \sim p_{data}(\pi)$ and $\pi,\mathbf{c} \sim p_{data}(\pi,\mathbf{c})$. We then train a conditional generator network $G(\mathbf{z},\mathbf{c})$ as below. The third term is an added regularizer that requires the generator to reconstruct the landing point of the policy that it just sampled. Here $G(\mathbf{z})_T$ means sample the policy and its target point and take only the target point term, and $G(\mathbf{z},\mathbf{c})_{\neg T}$ means the opposite.

$$\min_G \max_D V(G,D) = E_{\pi,\mathbf{c}\sim p_{data}(\pi,\mathbf{c})}[\log D(\pi,\mathbf{c})] + E_{\mathbf{z}\sim p(\mathbf{z}),\mathbf{c}\sim p_{data}(\mathbf{c})}[\log(1 - D(G(\mathbf{z},\mathbf{c})_{\neg T}))] + \|G(\mathbf{z},\mathbf{c})_T - \mathbf{c}\|_2] \tag{2}$$

*Policy Representation* We have applied our framework successfully to many different policy representations including sampling the RBF parameters of the forcing term of a DMP, but we found the following simple representation effective and easy to tune. For our Baxter robot arm, we represent the $\pi$ as a 15D vector defining a high-level open-loop controller in terms of the ball release time, and effector position and velocity at release. Specifically $\pi = [\boldsymbol{\theta}_{t_T}, \dot{\boldsymbol{\theta}}_{t_T}, t_T]$, where $\boldsymbol{\theta}_{t_T}$ and $\dot{\boldsymbol{\theta}}_{t_T}$ are 7D robot arm joint angles and joint velocities at launch time, and $t_T$ is the launch time. The goal condition $\mathbf{c}$ is a 2-dimensional Cartesian coordinate of the ball landing point. There are multiple launch configurations as described above, that result in the same landing point $\mathbf{c}$, and the trained GPN will sample this space of configurations.

*Policy Execution* With the policy definition above, samples from our GPN constitute a high-level action plan of how to launch the ball. To actually actuate this we map the high-level action into an open loop controller for low-level actuation via the following third-order polynomial function of time:

$$\boldsymbol{\theta}_{t_i} = \alpha_4 \left(\frac{t_i}{t_T}\right)^3 + \alpha_3 \left(\frac{t_i}{t_T}\right)^2 + \alpha_2 \left(\frac{t_i}{t_T}\right) + \alpha_1 \tag{3}$$

with $\dot{\boldsymbol{\theta}}_{t_i} = \frac{d\boldsymbol{\theta}_{t_i}}{dt_i}$ and $\ddot{\boldsymbol{\theta}}_{t_i} = \frac{d\boldsymbol{\theta}_{t_i}}{dt_i}$, and parameters:

$$v_{t_i} = \frac{1}{t_T}\left(3\alpha_4 \left(\frac{t_i}{t_T}\right)^2 + 2\alpha_3 \left(\frac{t_i}{t_T}\right) + \alpha_2\right),\ \alpha_1 = \boldsymbol{\theta}_{t_0},$$

$$\alpha_2 = \dot{\boldsymbol{\theta}}_{t_0} t_T, \quad \alpha_3 = 3\boldsymbol{\theta}_{t_T} - \dot{\boldsymbol{\theta}}_{t_T} t_T - 2\alpha_2 - 3\alpha_1,$$

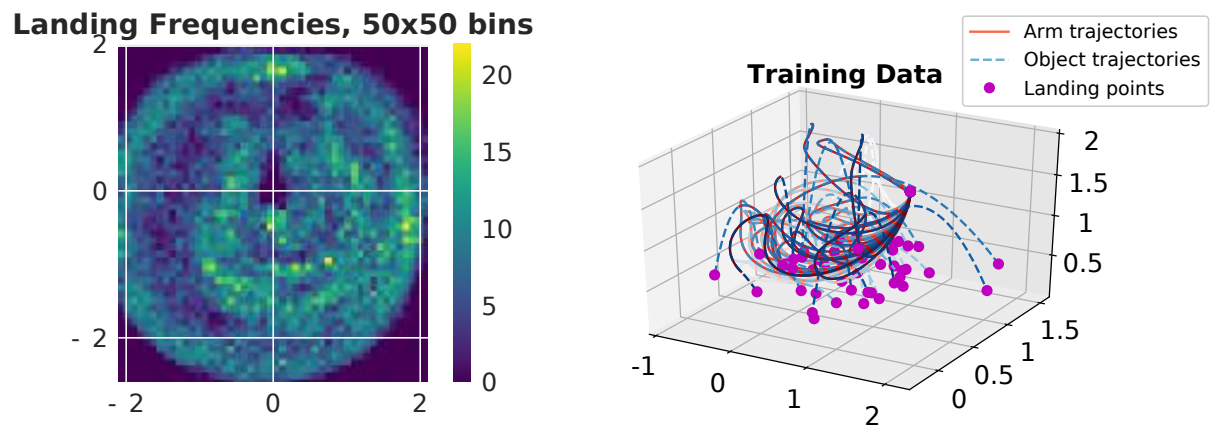

Fig. 1. Training data for throwing. Baxter robot is located at $(0,0)$. Left: Frequency of ball landing points at different floor positions ($50 \times 50$ bins). Right: Example trajectories where red line is the end-effector, blue lines are the ball in flight, and magenta points are the landing points on the floor.

$$\alpha_4 = \boldsymbol{\theta}_{t_T} - \alpha_1 - \alpha_2 - \alpha_3,$$

where $\boldsymbol{\theta}_{t_i}, \dot{\boldsymbol{\theta}}_{t_i}, \ddot{\boldsymbol{\theta}}_{t_i}$ are the positions, velocities and accelerations of joints at time $t$; $\boldsymbol{\theta}_{t_0}$ and $\dot{\boldsymbol{\theta}}_{t_0}$ are initial positions and velocities at time $t_0$ and $t_T$ is the time of launch. We assume the starting robot arm configuration is the same for each trial. Baxter is then actuated by sending the above joint position, velocity and acceleration plan to a ROS control node.

## IV. Experiments

### A. Training data and settings

We apply our GPN to enable a Baxter robot to robustly throw a ball in different environmental obstacle conditions.

*Training Data* We first describe the collection of data used to train our GPN. Training data is collected by using evolutionary Quality Diversity search [9] to find a set of genotypes (high level policies $\pi$) that have diverse behavior-space effects (e.g., arm trajectories and landing positions) when actuated. The data collection roughly follows the procedure described in [34], using a realistic Baxter simulation (Gazebo) to obtain around 15,000 throwing episodes. Each training episode records the arm trajectory, ball trajectory, and ball landing point. Since the Baxter arm has 7 joints, there are 15 parameters for any policy (Eq. 3) – position and velocity for each joint and the launch time. The training data is illustrated in Figure 1 in terms of a heat map of ball landings at different positions on the floor around the robot (left), and some example episodes' arm trajectory, ball trajectory and landing point (right).

*Evolutionary Baseline* We use QD search to generate training data for our GPN as described above, and will exploit our trained GPN to solve a robust throwing task later. A conventional evolutionary-style approach to exploiting this data for robust throwing would be to treat the dataset as a large library (repertoire), and then solve a new task by selection from the repertoire [3], [4], [5]. To throw to a specific target, the closest memorized landing point is recalled, and the associated policy is executed. If an environmental change (e.g., an obstacle) causes that known solution to fail, a lookup can be performed to find and execute some other policy with approximately the same landing point, but potentially different arm/ball trajectory. The problem is that this scales badly: although a large number of throwing episodes (15,000) covers the space of landing points reasonably well, it is not enough to cover many diverse ways to throw to each individual landing point. So the lookup-based approach may fail to effectively find diverse ways to throw to a specific point.

*Settings* Our GPN is built upon DCGAN framework [24], and has a 4-layer RELU-activated convolutional architecture that maps a 100-dimensional noise vector $\mathbf{z}$ to a 15-dimensional output vector representing $\pi$. It is trained using 15000 episodes of data using learning rate 0.0002 and 1000 epochs with batch size 250. We used 20 generator updates for each discriminator update. Both QD and GPN use the same policy representation $\pi$, and underlying actuation strategy.

### B. Experiment 1: Conditional throwing

*Setup* We aim to achieve robust throwing by learning to hit a target in diverse ways. We therefore first evaluate the ability of our GPN and QD alternative to: (i) accurately throw to a given position, and simultaneously (ii) find diverse ways of throwing to each position around the robot. For this purpose we grid the floor space around the robot into a 5×5 grid (25 target landing points). We experiment both in simulation (Gazebo Baxter) where we attempt to throw to each of those points 10 times, and then corroborate those results on the real Baxter where we throw to each coordinate 3 times, tracking the ball using an OptiTrack system. For each target point, we compute: **RMSE** between the target and actual landing point for all the trials; **Diversity** of the trials by taking the ball trajectory, computing equidistant waypoints along it, and then using these to compute a standard deviation of all trajectories towards a given target; **Harmonic Mean** between accuracy $(1-RMSE)$ and diversity (standard deviation) in order to provide a single quantitative measure of performance. We compare results from our GPN with the standard evolutionary strategy that treats the GPN-training set as a repertoire library (Sec. IV-A).

*Results* The results in Figure 2 plot the landing error and diversity metrics at each grid point on the floor around the robot. The first two columns compare QD/Library-based approach with our GPN in simulation; the third column evaluates our GPN results on the real Baxter robot. From the results of this experiment we make the following observations: (i) In general QD search has higher accuracy. This is expected as it is simply recalling previously memorized movements and replaying them exactly, which unsurprisingly leads to very similar outcomes, and hence high accuracy. In contrast our GPN is a predictive model that must infer the right policy to throw to any given target point, so its slightly lower accuracy is understandable. (ii) However, GPN has much higher diversity. It models the distribution of policies that throw to a conditioning target, and samples that distribution for each trial. (iii) Aggregating these metrics via harmonic mean, we see that our GPN performs favorably compared to QD. (iv) When executing our GPN-sampled controllers on the physical Baxter robot, the results are comparable to the simulated case (third vs second column). Note that the grey areas to the left of the map on the results of the real robot are because of walls in the physical Baxter environment preventing data collection there.

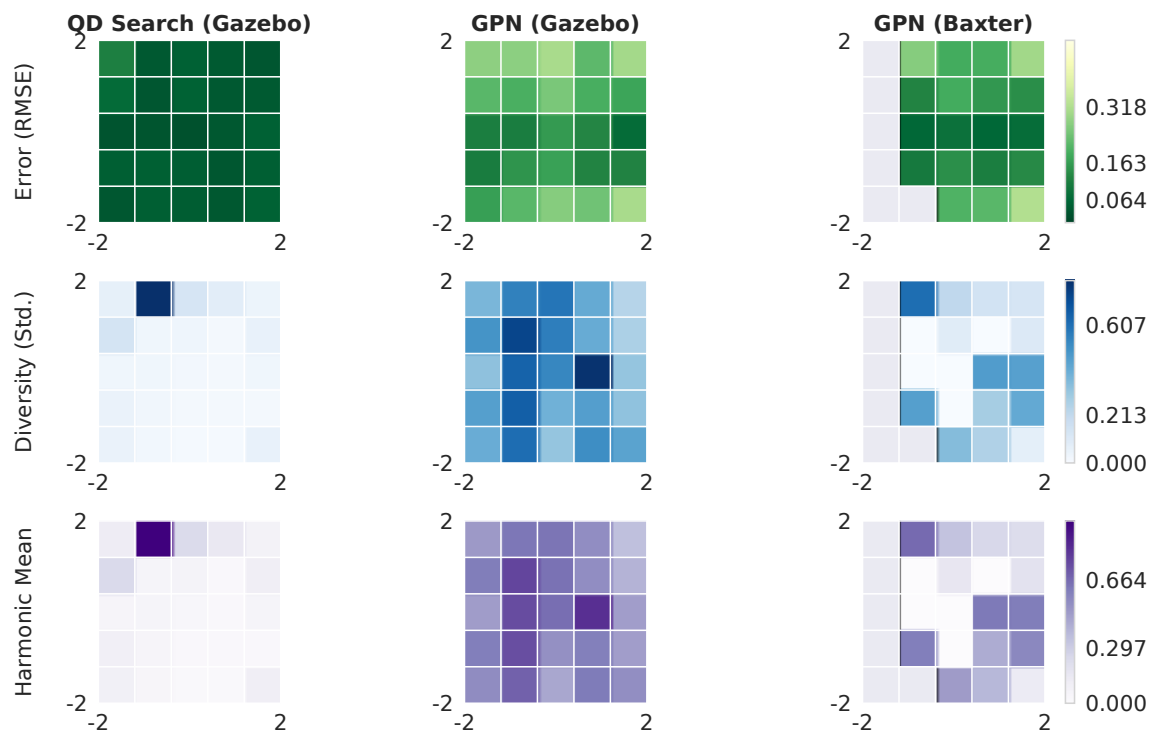

Fig. 2. Throwing to a 5×5 grid of points on the floor around the robot located at $(0,0)$ and facing right. Local error (top), diversity (middle), and their harmonic mean (bottom) when throwing by QD trajectories in simulation, GPN sampled controllers in simulation, GPN controllers on a real robot. GPN generally has higher diversity, and better overall performance (harmonic mean).

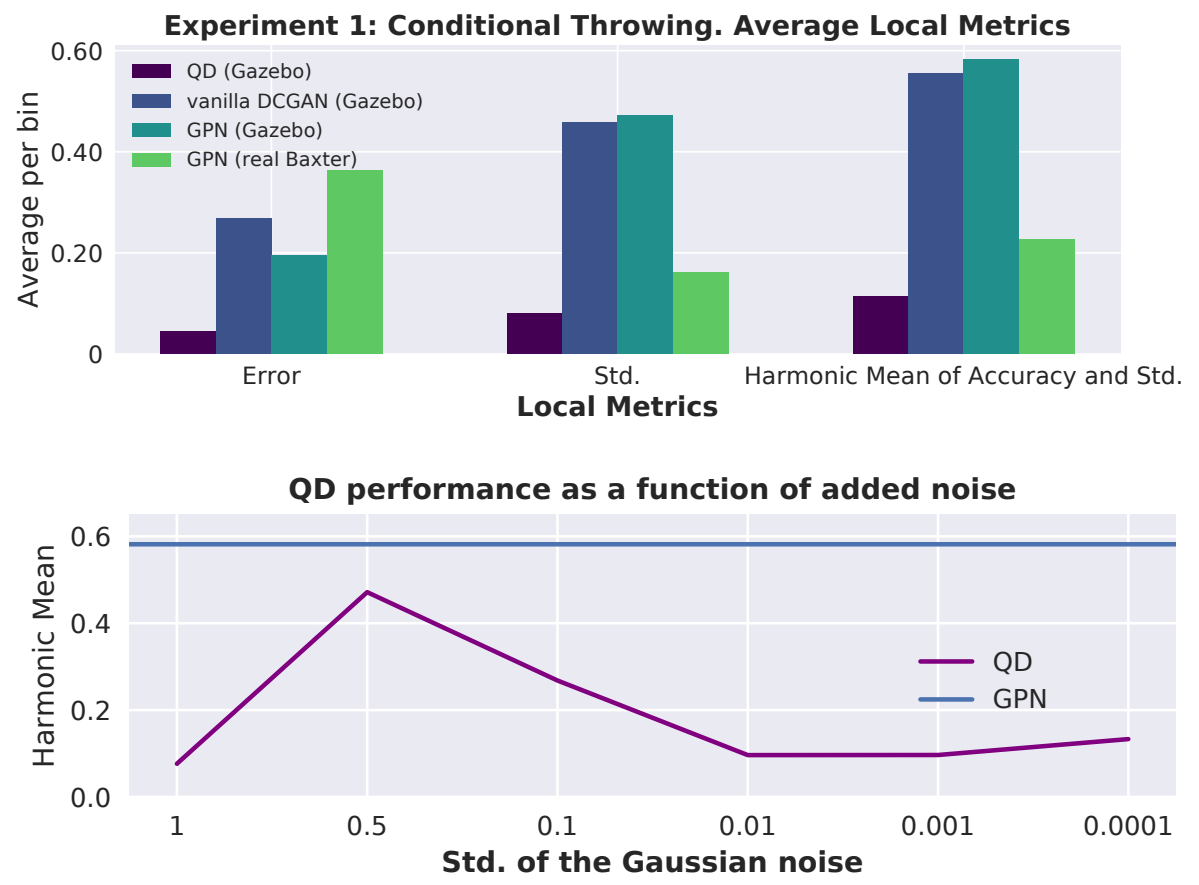

Fig. 3. Top: Summary statistics of throwing to all points on a $5\times5$ floor grid. The difference between harmonic means of the simulated data for QD and GPN is statistically significant according to unequal variances t-test with significance level $\alpha = 0.05 (p_{value} < 2.78 \cdot 10^{-12})$. Bottom: Comparison to QD with varying levels of added noise.

These results are summarized over all the spatial coordinates in Figure 3 (top), where GPN significantly outperforms QD in terms of harmonic mean. To be as fair as possible to the QD search alternative, we also considered boosting its diversity by adding Gaussian noise to the executed policies at each trial. The result in Figure 3 (bottom) shows that noise can improve QD performance. However it must be carefully tuned as too much noise quickly degrades QD's accuracy. Overall this result is understandable as uniformly adding noise to known throwing behaviors can quickly move off the manifold of good throwing policies. In contrast GPN learns the distribution over good throwing policies so it can sample novel throwing controllers from within that distribution. Finally we mention that, as per common safety practice, all our movement plans are checked for self-collision before execution. We also note that despite lacking a model of robot kinematics, the vast majority (98.4%) of the diverse plans generated by GPN are collision free. This indicates that GPN has also learned about the manifold of reasonable controllers in the sense of non-colliding as well as ability to hit a target.

### C. Experiment 2: Throwing with obstacles: GPN vs QD

*Setup* Our motivating scenario was to use the learned conditional distribution over controllers to achieve robust throwing in the presence of obstacles. In this experiment, we evaluate this quantitatively using Gazebo Baxter simulator. Specifically, we consider a $5\times5$ grid of floor targets as before, and we throw to each of these targets with 10 diverse sampled controllers as before. For each of those throws, we simulate obstacles and calculate whether an attempted throwing trajectory fails due to robot or ball collision with the obstacle. We consider a throwing trial as a success if the ball lands within radius $\tau$ of the intended target. Our metric is $SuccessesProportion(k,\tau)$: how many of the target coordinates does the ball hit successfully (within $\tau$ radius), at least $k$ out of 10 times. The idea is that even if obstacles block some particular throws, a model that can generate multiple diverse behaviors that all solve the task (i.e., throwing trajectories that hit the same target) should be able to find at least some (i.e., $k$) successful solutions. To systematically explore these issues we run the simulation for $k = 1\ldots9$, $\tau = 0, 0.1, \ldots, 1.0$, and repeat for different occlusion rates $= 1\%, \ldots, 8\%$. For simplicity we model occlusions as a randomly selected set of inaccessible floor areas, where the total proportion of blocked floor area is the specified occlusion rate. We compute results averaging over a $5\times5$ target grid, 10 throws per target, and 1000 random obstacle maps.

*Results* Figure 4 (right) shows the Gazebo simulation of Baxter attempting to throw to a specific target in the presence of randomly generated obstacles. Figure 4 (left) shows heat-maps of $SuccessesProportion(k,\tau = 0.2)$ for various occlusion rates and minimum hit requirements $k$. From these we can make the following observations: (i) Both QD and GPN methods have higher success rate in the easier bottom left (low occlusion, low hit ratio required for success), and vice-versa in the harder top right. (ii) The GPN result is much higher than that of QD for low $k$ values (e.g., $k = 1$). This means that the GPN can often find at least one way to hit the target, for this whole range of occlusion rates. (iii) At very high minimum hit (e.g., $k = 9$) QD performance is slightly better than GPN. This is because GPNs slightly lower accuracy means that it's rarely the case that as many as 9 out of 10 attempts hit the target. However, at this stringent hit rate requirement, we note that the success rate of QD in absolute terms is also very low (around 10%). Finally, Figure 4 (middle) shows the success rate averaged over occlusion rates as a function of different hit-radius requirements. We see that for a stringent accuracy requirement ($\tau < 0.1$), neither method succeeds. While for all larger values of $\tau$, GPN consistently outperforms QD in success rate. Overall the results validate the outcome we aimed to achieve: GPN can throw accurately enough that it often hits the target, but crucially it does so in diverse

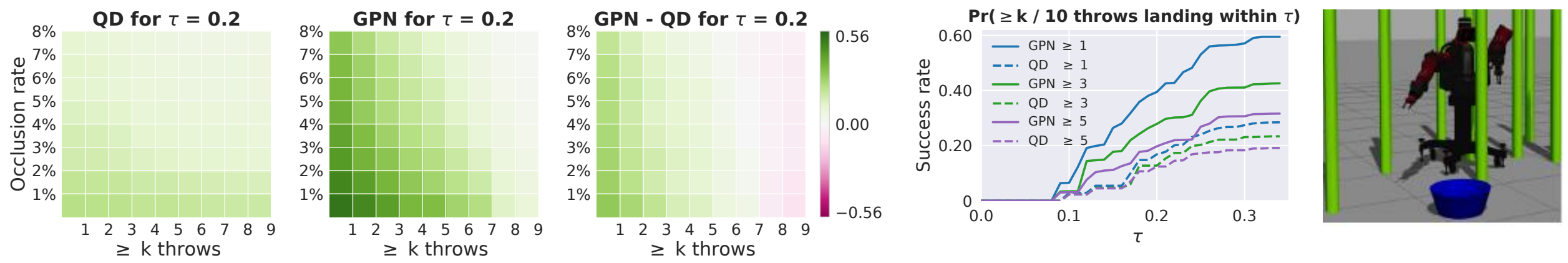

Fig. 4. Obstacle robustness of conditional throwing. Left: Heat maps illustrate the probability of at least $k$ of 10 throws landing within $\tau = 0.2$ of the target for different levels of occlusion. QD lookup method. Our GPN method. Difference between GPN and QD result. Middle: Success rate for varying accuracy requirements $\tau$. Right: Example of the random obstacle environment.

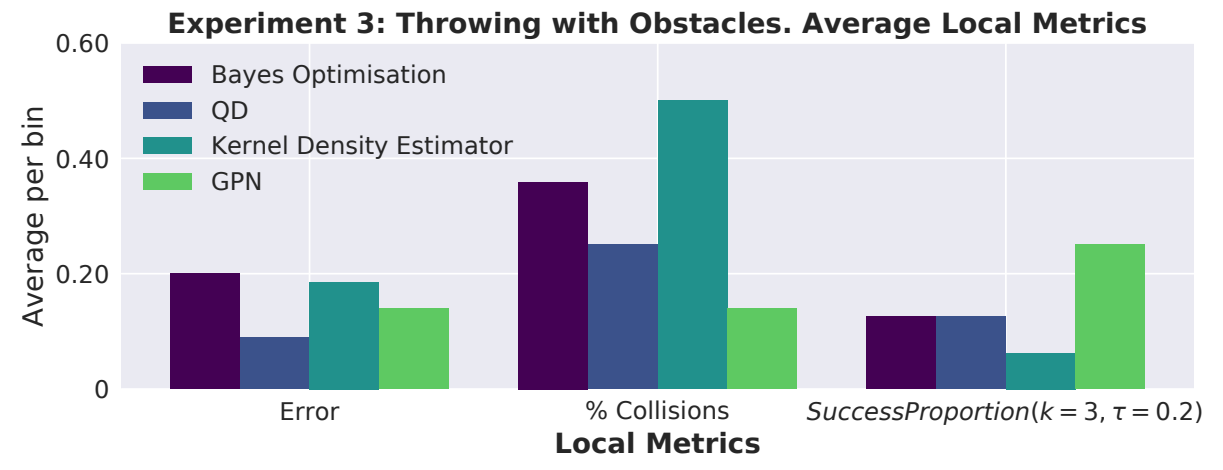

Fig. 5. Obstacle robust conditional throwing. Comparison of GPN, QD, Bayes Opt, and KDE.

enough ways that at least one way can usually be found to dodge any given obstacle configuration.

### *D. Experiment 3: Throwing with obstacles comparison*

*Setup* In this experiment, we compare two further alternative approaches to obstacle-robust throwing. KDE: As a non-parametric alternative to our GPN, we define a target-conditional Kernel Density Estimation [35] model over the same QD-based training set used by our GPN, so $p(\pi|\mathbf{c})$ is a Gaussian mixture model. We can then sample this mixture instead of our GPN. BayesOpt: Bayesian Optimisation [36] is an established approach to adaptive behaviours in robotics [17]. We use the the best QD-trajectory as the starting condition, and then perform Bayesian optimization for 10 trials for direct comparison to the diversity based models.

*Results* We use a simulated setup where there is a wall randomly placed between Baxter and the target (see supplementary video and Section IV-E). The results in Figure 5 average over four random goal/wall positions and compare the different methods in terms of error, diversity, collision rate and $SuccessesProportion(k = 3, \tau = 0.2)$. We see that while GPN is not the most accurate model, its success rate is best overall. This is because (i) it has good diversity enabling it to dodge obstacles more often, (ii) it has learned the manifold of reasonable policies, so it usually also avoids self-colliding or unsafe movements. In contrast, KDE and BayesOpt often generate self-colliding movements or hit the obstacle. KDE suffers from being an inefficient/inaccurate model of relatively high-dimensional (15D) policies. BayesOpt purposefully adapts the movement to avoid the obstacle, but cannot succeed in the relatively small number (10) of available trials.

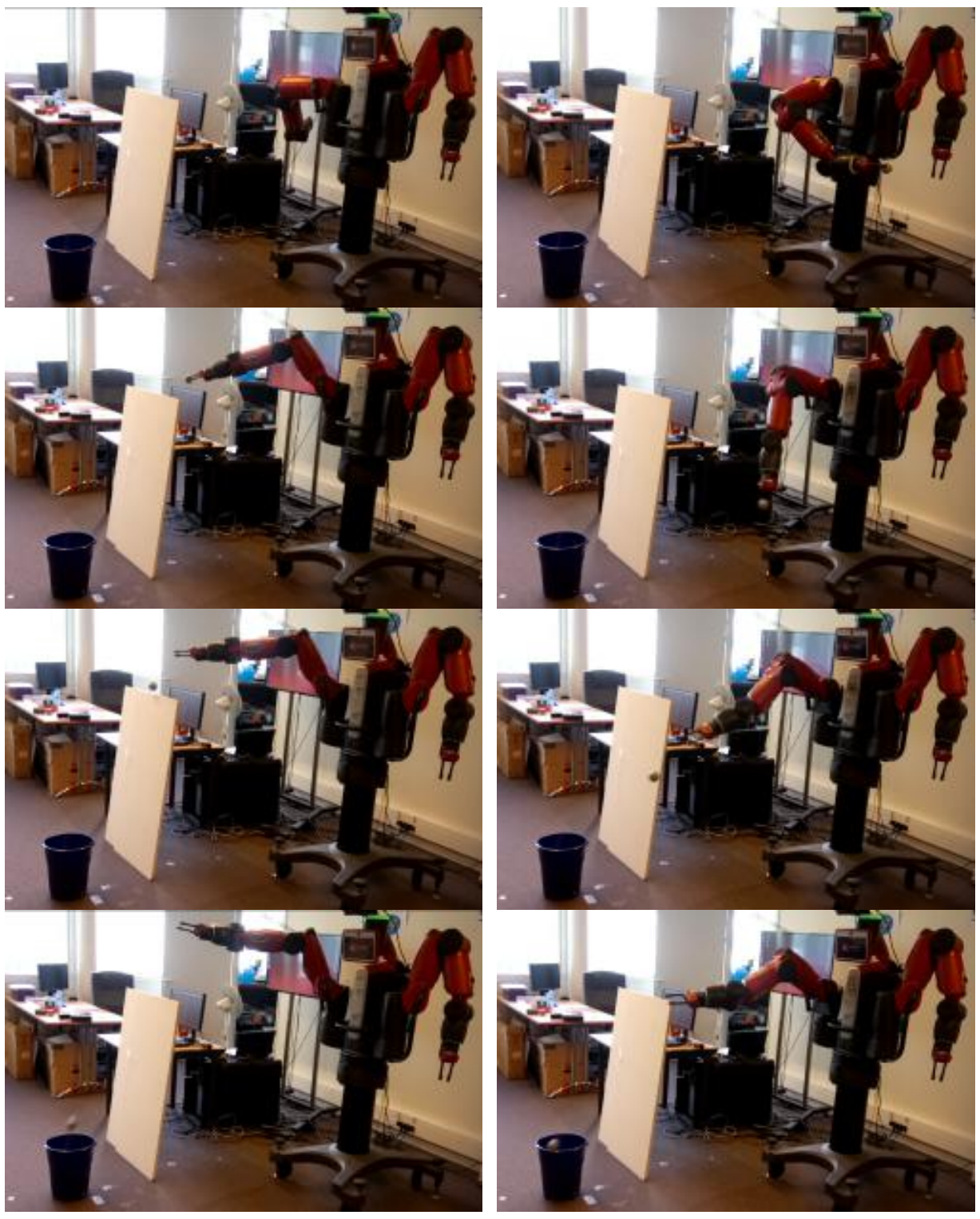

Fig. 6. Two examples of obstacle robust throwing behaviors obtained by sampling our learned distribution over policies. The GPN is conditioned on the target location, and samples controllers for throwing there until samples are drawn that generate neither robot nor ball collisions.

### *E. Experiment 4: Physical Baxter robust conditional throwing with obstacles*

We finally demonstrate our framework enabling physical Baxter robot to avoid an obstacle when throwing at a specific target. The required target coordinate is given in the previous experiments, but many typical routes to throwing to this target are blocked by the obstacle. In this case the obstacle's position means that a successful throw needs to go over and above the obstacle, or (slightly awkwardly for a right-handed throw) to the left of the obstacle. However for this configuration, *every trajectory in the training library collides with the obstacle*. Therefore the standard QD/lookup-based approach fails to find any successful solution. To test our method, we sample the conditional GPN ten times to generate ten diverse controllers that should throw to the required point. We simulate them to check for collisions with the obstacle, and found three of these avoided collision and landed into the basket in simulation. This validates that the GPN has indeed learned to generalize and samples novel behaviors. Figure 6 shows the successful execution of two of these controllers. We can see that the ability to generate diverse trajectories enables the robot to successfully hit the target with the ball while dodging the obstacle. For a video of this experiment please refer to the supplementary material.[1]

## V. Conclusion

We introduced the idea of generative policy networks, for defining a generative model over policies. We showed that our generative policy network provides a way to compactly encode a large set of known behaviors, and that sampling the GPN provides a way to draw unlimited novel controllers that are related-to but different-from known training behaviors. We showed how to apply this novel idea to robustly solving tasks, with a specific example in the form of obstacle-robust throwing. In future work we intend to explore applying the proposed generative policy network framework for generating closed-loop rather than open-loop controllers, application to different kinds of tasks besides throwing, and exploiting the generator online for optimal training data collection rather than relying on a fixed training set.

## Acknowledgement

This work is supported by the DREAM project through the European Unions Horizon 2020 research and innovation under grant agreement No 640891.

[1] https://youtu.be/2LCnaa89erM